\documentclass{article}

 \usepackage[final]{nips_2017}

\usepackage[utf8]{inputenc} 
\usepackage[T1]{fontenc}    
\usepackage{hyperref}       
\usepackage{url}            
\usepackage{booktabs}       
\usepackage{amsfonts}       
\usepackage{nicefrac}       
\usepackage{microtype}      

\usepackage{subcaption}
\usepackage{mathtools}

\usepackage{amsthm}
\usepackage{wrapfig}

\theoremstyle{definition}

\author{
Ashley D. Edwards, Srijan Sood, Charles L. Isbell Jr. \\
College of Computing\\
Georgia Institute of Technology\\
\texttt{$\{$aedwards8, srijansood$\}$@gatech.edu, isbell@cc.gatech.edu}
}

\title{Cross-Domain Perceptual Reward Functions}

\begin{document}
\maketitle

\begin{abstract}
In reinforcement learning, we often define goals by specifying rewards within desirable states. One problem with this approach is that we typically need to redefine the rewards each time the goal changes, which often requires some understanding of the solution in the agent's environment. When humans are learning to complete tasks, we regularly utilize alternative sources that guide our understanding of the problem. Such task representations allow one to specify goals on their own terms, thus providing specifications that can be appropriately interpreted across various environments. This motivates our own work, in which we represent goals in environments that are different from the agent's. We introduce Cross-Domain Perceptual Reward (CDPR) functions, learned rewards that represent the visual similarity between an agent's state and a cross-domain goal image. We report results for learning the CDPRs with a deep neural network and using them to solve two tasks with deep reinforcement learning.
\end{abstract}

\section{Introduction}
Rewards often act as the sole feedback for Reinforcement Learning (RL) problems. This signal is surprisingly powerful---it can motivate agents to solve tasks without any further guidance for how to accomplish them. Nevertheless, rewards do not come for free, and are typically hand-engineered for each problem. Furthermore, they must traditionally be defined in terms of the~\emph{agent's} environment, which may be difficult or tedious. Such reward specifications can be practical for single-goal problems, yet challenging in environments consisting of multiple goal configurations. We may be willing, for example, to specify rewards based on the numerous cuts, holes, and bindings that are necessary to construct a table, but developing rewards for each type of furniture could be intractable.

Learning from demonstration approaches, such as inverse reinforcement learning and imitation learning~\cite{argall2009survey}, aim to alleviate the
engineering effort by utilizing examples of how a task should be solved. These techniques can be valuable when training an agent~\emph{how} to complete a task, but are not always necessary when we wish to specify~\emph{what} the task is. Alternatively, goals may be specified through target images, but an underlying concern remains with this and each of the former approaches. That is, we must represent the problem in the agent's environment.

One common technique humans use when approaching problems is to study replicas of the solved instance. We may, for example, attempt to build a table by examining an image of one that is already configured. This is analogous in RL terms to specifying a reward based on the agent’s environment, either through the parameters of the agent's state or target images. But we may also use other resources to guide our solutions---written instructions, spoken words, diagrams, etc. We aim to allow specifying tasks in similar manners, across domains, through cross-domain goal instances that are defined in environments outside of the agent's. Rather than hand-specifying rewards based on internal parameters of the agent, or providing target images from the agent's environment, our approach allows specifying goals in surrogate environments, where one may more easily find or construct solutions.

We introduce Cross-Domain Perceptual Reward (CDPR) functions, which produce rewards that represent deeply learned similarities between a cross-domain goal image and states from an agent's environment. ~\textbf{We empirically demonstrate that CDPRs are a~\textit{general} approach for providing~\textit{accurate} rewards between these cross-domain and intra-domain representations.}

We provide evidence of generality by showing that CDPRs can be used without modification across multiple goal instantiations. We demonstrate accuracy by introducing a goal retrieval metric and additionally showing that CDPRs can successfully train an agent to complete a task. We describe two novel tasks with two distinct cross-domain goal specifications for each. We show that by using CDPRs as a replacement for hand-specified reward functions, we can successfully train an agent to solve these tasks with deep reinforcement learning, with the added benefit that we can specify the goals on our own terms.

The rest of the paper is organized as follows. We begin by providing the background for our approach in Section~\ref{sec:background}. In Section~\ref{sec:related} we discuss related work and then in Section~\ref{sec:approach} we describe a deep neural network architecture that learns the CDPRs. We provide results in Section~\ref{sec:experiments} followed by the conclusion in Section~\ref{sec:conclusion}.

\section{Background}
\label{sec:background}
Reinforcement Learning (RL) problems are traditionally specified by a Markov Decision Process $\langle S, A, P, R \rangle$~\cite{suttonbarto}. The set $S$ consists of states $s \in S$ that correspond to the agent's environment. An agent takes actions $a \in A$ and receives rewards $r \in R(s)$ that depend on the current state. In general, an agent's task is defined solely by the rewards specified in the MDP. Each time the task changes, the reward function must be redefined. The transition function $P(s, a, s')$ represents the probability that the agent will land in state $s'$ after taking action $a$ in state $s$. A policy $\pi(s,a)$ represents the probability of taking action $a$ in state $s$. We typically are interested policies that maximize the long-term expected reward over time. In our problem, we are not given rewards as inputs to the MDP. Rather, we aim to use deep learning to learn a reward function that reflects the similarity between an image of the agent's state and an image from another environment. There has been a large body of work in deep learning for finding similarities across domains, generally applied to image classification, generation, and retrieval problems~(e.g. \cite{DBLP:journals/corr/BousmalisTSKE16, chopra2005learning, harwath2017learning, DBLP:journals/corr/IsolaZZE16, DBLP:journals/corr/LarsenSW15}). Our work explores the use of such techniques for cross-domain goal specifications.

\section{Related work}
\label{sec:related}
We now describe several approaches for representing goals in RL. Throughout the literature, we have found that goals are often defined in terms of the agent’s environment. Our own approach introduces a general mechanism for specifying goals on our own terms. We discuss two broad techniques for representing goals, either by explicitly instantiating them through engineering or direct feedback, or implicitly expressing them through demonstrations.

\subsection{Explicit goal instantiations}
Arguably, one of the most common approaches for representing goals for RL problems is to provide hand-specified rewards that are based on internal parameters of an agent's environment. There is a wealth of literature that describes such task-specific rewards, from those that indicate when an agent has reached a desired location~\cite{sutton1990integrated}, to more complex ones that can be used as feedback for training neural networks~\cite{zoph2016neural}. Rewards are largely where we instill domain knowledge, and so are inherently specialized. Even if a reward is commonly used for multiple problems in the same domain, it often remains a function of the configurations of each agent in the environment.

For problems consisting of visual inputs, a more general solution for representing goals is to use target images. With this approach, rewards are based on pixels alone, thus allowing one to abstract away the task specification from the parameters of the agent’s configuration. This approach is similar to visuo-servoing, which uses target images to guide robots to some goal~\cite{hutchinson1996tutorial}. Such solutions often require extracting known objects from camera images. A more recent RL approach learns the relevant features from the agent’s state space~\cite{finn2016deep}. However, this method still requires defining the goal in terms of the agent's environment. Another approach uses motion templates to transform the visually dissimilar motions of humans and a simulated robot into a similar structure, but this correspondence is hand-specified~\cite{edwards2016perceptual}. Our approach aims to learn such correspondences.

When multiple goals or MDP settings exist, it may be necessary to produce more general solutions. Early work aimed to find rewards that generalized across multiple environments~\cite{singh2009rewards}. Recently Finn et al.~\cite{finn2016generalizing} described an approach that allowed generalizing skills learned from ``labeled'' MDPs with known reward functions to similar ``unlabeled'' MDPs with unspecified reward functions. Our work also uses a semi-supervised approach for learning reward functions, but we do not assume the labeled and unlabeled distributions are similar. Finally, we have seen approaches that aim learn a value function that generalizes over states and goals, but these approaches also assume that the two representations share a similar structure~\cite{schaul2015universal, zhu2016target}.

Specifying the goal itself may not always be difficult, but training an agent to actually reach the goal may be. Training in simulation and then transferring the knowledge to the real-world can often be more efficient than training there directly (e.g. ~\cite{rusu2016sim,sadeghi2016cad,tobin2017domain, tzeng2016adapting, zhang2015towards}). Such specifications may also be considered cross-domain, although the motivation for these works is different from our own. In particular, such ``sim-to-real'' approaches tend to focus on transferring across separate realizations of similar domains.

Many methods utilize human feedback to guide an agent's behavior. For example, some approaches allow humans to provide rewards to agents in real-time~\cite{isbell2001social, knox2015framing, thomaz2008teachable}. Alternatively, policy shaping takes a more hands-off approach, and provides policy ``advice'', as opposed to explicit rewards~\cite{griffith2013policy}. Another approach aims to interactively teach an agent goals~\cite{kirk2016learning}. Finally, rather than providing direct target images for where an object should be, Finn et al.~\cite{finn2016deep} introduced an approach that allowed humans to select the pixel locations for where a robot should move an object.

\subsection{Implicit goal instantiations}
\label{sec:demonstration}
Learning from Demonstration (LfD) is often used when specifying a goal is difficult, or when a problem is too challenging for an agent to solve on its own. Our approach differs from LfD, as we do not aim to portray how to solve a task, only what the task is. Nevertheless, we do share some motivation with these works, which we now discuss.

Inverse RL aims to infer a reward function from expert demonstrations~\cite{abbeel2004apprenticeship}. This approach typically requires multiple demonstrations for a single task. We also aim to learn a reward function in our approach, but we use a single sample to represent the goal.  A similar approach is imitation learning, which aims to learn a policy directly from demonstrations~\cite{schaal1999imitation}. Both of these techniques often assume that the student, i.e., agent, and the teacher, or demonstrator, share a common environment. That is, the observed states and actions are shared between the teacher and student. When the observations are dissimilar, a correspondence problem must be solved~\cite{argall2009survey}. These correspondences are often hand-specified, and may require equipment such as sensors on the teacher’s body. Our own approach aims to learn such correspondences between different representations.

Recent works have proposed solving the correspondence problem automatically. For example, one approach used deep learning to train an agent that had a first-person perspective through third-person samples~\cite{stadie2017third}.  Another somewhat related RL approach is transfer learning, which aims to transfer learned behaviors from one domain to another~\cite{taylor2009transfer}, for example by initializing the parameters of policies in unsolved tasks~\cite{ammar2015unsupervised}, or by transferring skills across untrained robots~\cite{devin2016learning}. Our own approach does not assume we have examples of desired behaviors.

Finally, there has been a breadth of recent work that aims to specify goals in different modalities. Such approaches have similar a motivation to our own. That is, they aim to provide goal specifications in more natural settings than the agent’s. For example, we have seen sketches of maps used for representing desired trajectories in navigational tasks~\cite{boniardiautonomous, skubic2001extracting}, correspondences learned between words and robotic actions~\cite{stramandinoli2011towards}, rewards based on the touch of a handshake~\cite{qureshi2016robot}, and value functions learned from facial expressions~\cite{veeriah2016face}. Recently, an approach learned correspondences between written instructions and visual states in Atari games~\cite{kaplan2017beating}. This approach is more similar to our own as it learns the correspondences between two different domain representations. It is clear that there is much momentum in this area of research. Still, we have not yet seen an approach that learns a visual correspondence for cross-domain goal representations.

\section{Approach}
\label{sec:approach}
\begin{figure}[t]
\centering
\includegraphics[width=.8\textwidth]{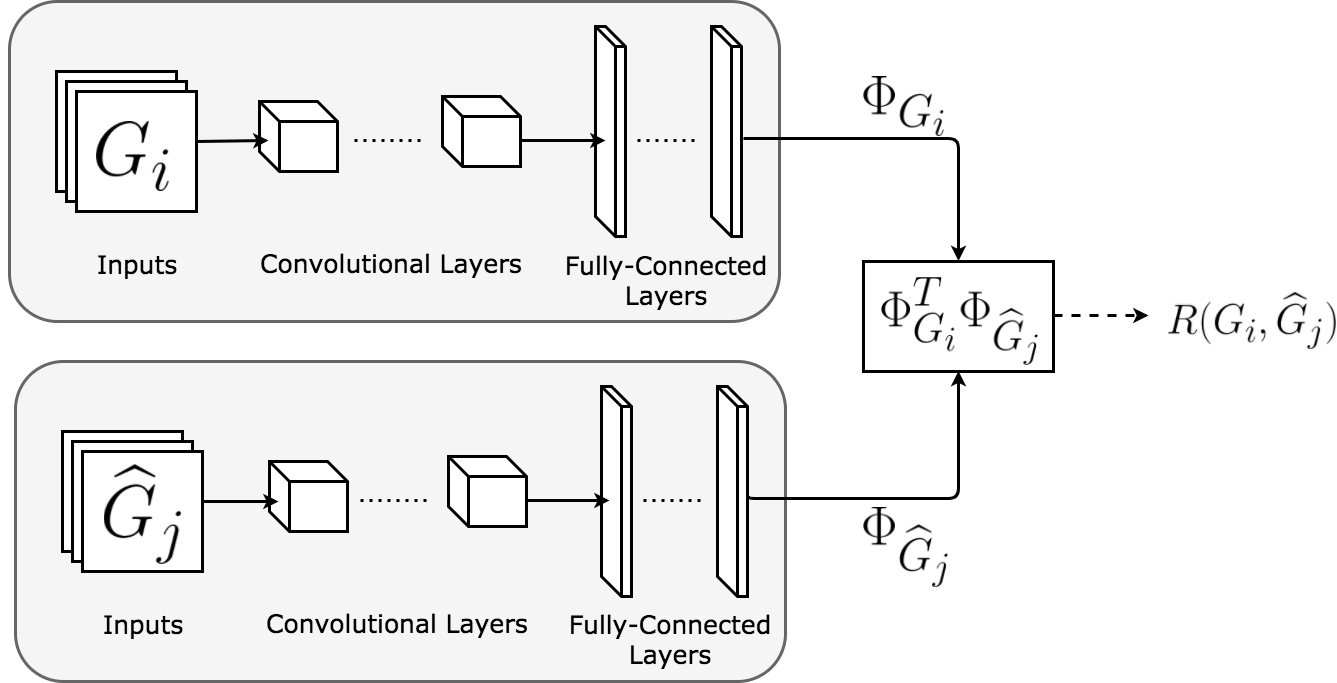}
\vspace{.6cm}
\caption{Deep neural network used for learning Cross-Domain Perceptual Reward (CDPR) functions. Two separate, though identical, networks encode features for the agent states $G_i$ and cross-domain goals $\widehat{G}_i$, respectively. Each network consists of a set of convolutional layers followed by a set of fully-connected layers. The exact components of each layer are task-specific. The dot product between the outputs of each encoding layer represents the CDPR, $R(G_i, \widehat{G}_j)$.}
\label{fig:network}
\end{figure}
We now formalize our approach for developing Cross-Domain Perceptual Reward (CDPR) functions. We focus on solving RL problems consisting of MDPs with visual states and unspecified reward functions. We are interested in problems where an agent has a task with multiple potential goal configurations. An agent's overall task may be to build furniture, for example, but its goal could be to build a chair, or table, or bench. Put succinctly, each goal is a specific instance of a task. Typically, an agent's task is defined by a reward function. This remains true for our approach as well, but now we require the reward to be a function of both the state and a goal. Therefore, for a single task, this~\emph{general} reward function remains fixed when the goal changes---only the inputs vary. We now describe how we develop such rewards, or CDPRs.

We aim to specify goals through images obtained from alternative environments. Our work is inspired by an approach that uses deep learning to find cross-domain similarities between images and their corresponding spoken captions, which are represented through spectrograms~\cite{harwath2017learning}. We hypothesize that this approach can be used in RL to represent cross-domain goals. The joint audio-visual approach used a network tailored to the spoken captions, but we aim to use a more general network that can accept any image as input. A diagram of our network is shown in Figure~\ref{fig:network}. The inputs to the network are states from the agent's environment, $G_i$, and cross-domain goal images, $\widehat{G}_j$. We use this notation for the agent's state from now on, as we consider each state image to be a potential goal image. The role of the CDPR is to determine this by outputting similarities between the two.

We construct two separate networks with identical architectures to encode the intra-domain features, $\Phi_{G_i}$, and the cross-domain features, $\Phi_{\widehat{G}_j}$, respectively. The CDPR is obtained by taking the dot product of these two outputs:
\begin{equation}
    R(G_i, \widehat{G}_j) = \Phi^T_{G_i} \Phi_{\widehat{G}_j}
\end{equation}
This reward then acts as the similarity between an agent's state and a cross-domain goal.

We train the network with a semi-supervised approach. Assume that for a given task, that we have $k$ pairs consisting of an intra-domain goal specification and a corresponding cross-domain goal: $\{(G_1, \widehat{G}_1), \dots, (G_k, \widehat{G}_k$)\}. After training, we only use cross-domain images to instantiate goals. We aim to learn a reward function that makes $R(G_i, \widehat{G}_j)$ large when $i = j$ and small otherwise. As such, we use the loss function from the previously described approach, which optimizes for exactly this target, modified appropriately for learning reward functions. Given the parameters $\theta$ of the network described in Figure~\ref{fig:network}, the loss can be formulated as:
\begin{equation}
\mathcal{L(\theta)} = \max\left(0, R_{\theta}(G_j, \widehat{G}_i) - R_{\theta}(G_i, \widehat{G}_i) + 1\right) + \max\left(0, R_{\theta}(G_i, \widehat{G}_j) - R_{\theta}(G_i, \widehat{G}_i) + 1\right)
\label{loss}
\end{equation}
We use gradient descent to optimize this loss function. For each training batch, we randomly sample two goal pairs, $(G_i, \widehat{G}_i)$ and $(G_j, \widehat{G}_j)$. The loss aims to make the reward for the matching pair, $R(G_i, \widehat{G}_i)$, be larger than the rewards for mismatched pairs, $R(G_j, \widehat{G}_i)$ and $R(G_i, \widehat{G}_j)$. Otherwise, the outputs for the loss function will be greater than $0$ and so the gradients will be penalized. Once we obtain the CDPRs, we can use them to instantiate the reward function for the MDP, and use standard RL approaches to solve the task. We should note that while the CDPR remains fixed across goal specifications, it is necessary to learn a new CDPR for each type of cross-domain goal representation.
\section{Experiments and results}
\label{sec:experiments}
\begin{figure}
\centering
\begin{subfigure}[b]{6.5cm}
\fbox{\includegraphics[height=4.2cm]{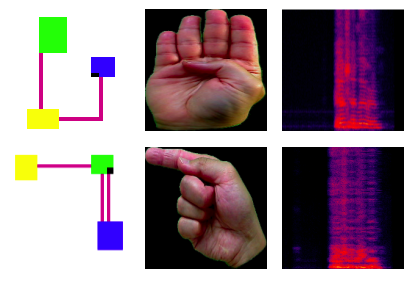}}
\caption{Maze Domain}
\label{fig:MazeDomain}
\end{subfigure}
\hspace{.35cm}
\begin{subfigure}[b]{6.5cm}
\fbox{\includegraphics[height=4.2cm]{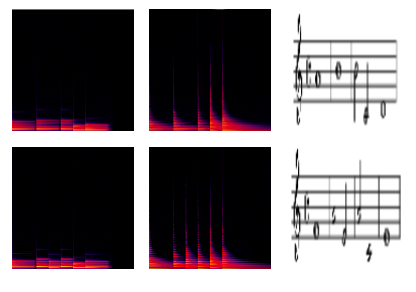}}
\caption{Music Domain}
\label{fig:MusicDomain}
\end{subfigure}
\caption{Domains used in RL experiments. In the Maze domain, the agent's task is to navigate to a specific room. The goal specifies the room the agent needs to navigate to. For the cross-domain goals, we used a sign language handshape of the first letter of the desired room's color, and a spectrogram of a spoken command, ``Go to the [color] room.'' In the Music domain, the agent's task is to play songs by selecting synthesized piano notes. For the cross domain goals, we used a spectrogram of a guitar that has played the desired song, and the sheet music for the song. The leftmost columns for Figures~\ref{fig:MazeDomain} and~\ref{fig:MusicDomain} represent two intra-domain goal representations for each task, while the remaining two are the cross-domain goal representations. The top and bottom rows for both  domains are referred to as Task $1$ and Task $2$ in our experiments.}\label{fig:domains}
\end{figure}
\begin{table}[htb]
\centering
\caption{Goal retrieval accuracy. We randomly obtained $5000$ goal pairs that were not seen while training and obtained the GRA for each cross-domain goal representation. We also include results for random retrieval and the GRA for the state encoding, but these methods do not use cross-domain specifications. As such, the handshape/speech and guitar/sheet music columns refer to the same results for the Maze and Music tasks, respectively.}
\bigskip
    \begin{tabular}{ | l | l | l | | l | p{2cm} |}
    \hline
    \textbf{Reward Function} & \textbf{Hand shape} & \textbf{Speech} & \textbf{Guitar} & \textbf{Sheet Music} \\ \hline
    Random & 0.342 & 0.342 & 0.002 & 0.002 \\ \hline
    State Encoding & 1.0 & 1.0 & 1.0 & 1.0 \\ \hline
    CDPR (Ours) & 0.98 & 0.986 & 0.904 & 0.888 \\
    \hline
    \end{tabular}
    \label{table:retrieval}
\end{table}

We aim to demonstrate the generality and accuracy of CDPRs. We compare against a standard hand-specified reward, and a state encoding reward that learns features in the agent's domain and represents the goal as an intra-domain target image. This reward is represented by the negative euclidean distance between the agent's state and the intra-domain goal representation. For each of the tasks, we automatically generated the cross-domain goal representations, as we aim to evaluate first if our approach works. Future work would entail using hand-specified goal specifications. We now describe the tasks we used to evaluate our approach.

We now describe a metric for measuring accuracy in reward functions. While evaluating with RL can suffice, it may be intractable to iterate over many goal instantiations. Therefore, we introduce the Goal Retrieval Accuracy (GRA) metric, which is motivated by image retrieval approaches. Given a set of states that were not seen during training time, we aim to find if the accurate cross-domain goal can be retrieved. In particular, this metric measures if the highest reward given to an unseen state matches the correct cross-domain goal. This measure gives an indication for how accurate the CDPRs are, and additionally indicates how well the learned rewards can generalize across multiple goal specifications.

We additionally measure how well the CDPRs work with RL. To further measure accuracy and generality, we used two unseen goal instantiations for each task, as shown in Figure~\ref{fig:domains}. We used Deep RL to solve the tasks with the architecture described in the paper~\cite{mnih2013playing}. We trained the network with an Adam optimizer with an initial learning rate of $10^{-4}$ and a batch size of $32$. To evaluate the agent's performance, we ran its learned policy on the task every $100$ episodes and measure a task-dependent evaluation accuracy.

\subsection{Maze task}
The first domain we evaluate our approach on is a Maze, as shown in Fig~\ref{fig:MazeDomain}. We randomly generated mazes consisting of 1-3 green, blue, or yellow colored rooms. The agent's actions are to move up, down, left or right, and its task is to navigate the maze. Goals designate the specific room the agent needs to reach. In order to specify the desired room using standard RL approaches, we would need to know its coordinates. Rather, we utilize two cross-domain goal representations: a sign language handshape indicating the first letter of the desired room color, and a spectrogram of a spoken command stating ``Go to the [color] room.'' We use a dataset of sign language gestures from~\cite{BarczakEt11_C} for the handshape specification. For the speech specification, we recorded one sample of the spoken command for each colored room, and then varied the pitch to produce multiple samples. We set the reward for the standard approach as $1$ if the agent is located in the desired room and $0$ otherwise. The agent's episode resets after $1000$ steps, as a terminal state requires knowledge of the goal location. During training, the agent could be initialized in any location on the map. During evaluation, it is initialized in the green room in the top maze in Figure~\ref{fig:MazeDomain} and in the yellow room in the bottom maze.

In order to train the CDPRs for the handshape and the speech specifications, we used a momentum optimizer with an initial learning rate of $10^{-4}$ and momentum value of $.9$. We used a batch size of $32$. The embedding networks for the CDPR have the following architecture: Conv1 $\rightarrow$ maxpool $\rightarrow$ Conv2 $\rightarrow$ maxpool $\rightarrow$ FC1 $\rightarrow$ FC2, where Conv1 consists of 64 11x11 filters with stride 4, Conv2 consists of 192 5x5 filters with stride 2, FC1 outputs 400 features, and FC2 outputs 100. Each maxpool consists of a 3x3 filter with stride 2. Conv1, Conv2, and FC1 are each followed by batch normalization~\cite{ioffe2015batch} and then ELU~\cite{clevert2015fast}.

\subsection{Music playing task}
The next domain we evaluate our approach on allows an agent to play synthesized piano notes by selecting keys and note durations. The agent can play $7$ keys on a single scale of a-g, and each key can be a whole or half note. When the agent selects a note, a .wav file is generated that gets converted into a spectrogram, which represents its state. The task is for the agent to play a song, and the goals designate which song to play. In order to specify the desired song, we would need to know how to play the notes on the piano. Rather, we again use two cross-domain goal representations: a spectrogram of the desired song played on a synthesized guitar and the sheet music of the song. To obtain the samples of both specifications, we randomly generated notes and automatically created the sheet music and spectrograms for the guitar. We set the reward for the standard approach as the total percentage of notes the agent has played correctly. The agent's episode resets after it plays notes for four complete bars. To avoid positive loops with the standard specification and CDPRs, the agent only receives a reward when it reaches this terminal state. During training and evaluation, the agent is initialized without any notes played.

In order to train the CDPRs for this task, we used an Adam optimizer~\cite{kingma2014adam} with an initial learning rate of $10^{-4}$. We used a batch size of $32$. The embedding networks for the CDPR have the following architecture: Conv1 $\rightarrow$ maxpool $\rightarrow$ Conv2 $\rightarrow$ maxpool $\rightarrow$ FC1, where Conv1 consists of 32 11x11 filters with stride 4, Conv2 consists of 64 5x5 filters with stride 2, and FC1 outputs 2048 features. Each maxpool consists of a 3x3 filter with stride 2. Conv1 and Conv2 are both followed by batch normalization and then ELU.

\subsection{Results}
\begin{figure}[htb]
\centering
\begin{subfigure}{.325\linewidth}
{\includegraphics[height=2.1cm]{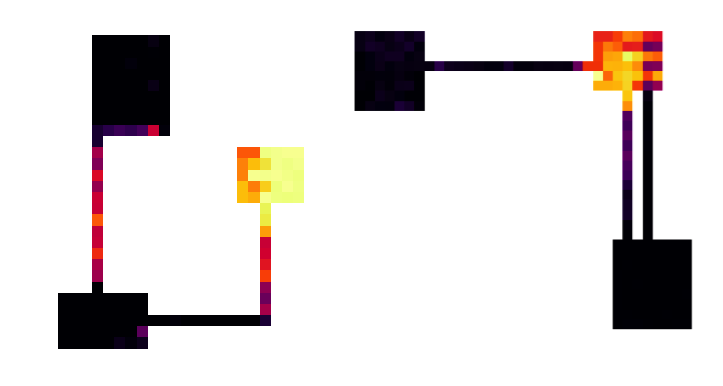}}
\caption{Handshape CDPR}
\label{Fig:handshapeCDPR}
\end{subfigure}
\begin{subfigure}{.325\linewidth}
{\includegraphics[height=2.1cm]{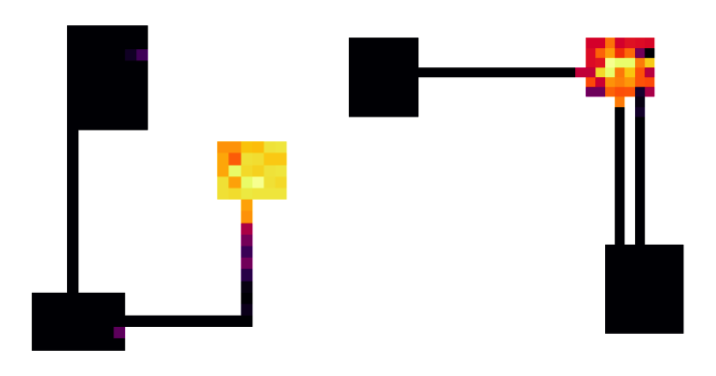}}
\caption{Speech CDPR}
\label{Fig:speechCDPR}
\end{subfigure}
\begin{subfigure}{.325\linewidth}
{\includegraphics[height=2.1cm]{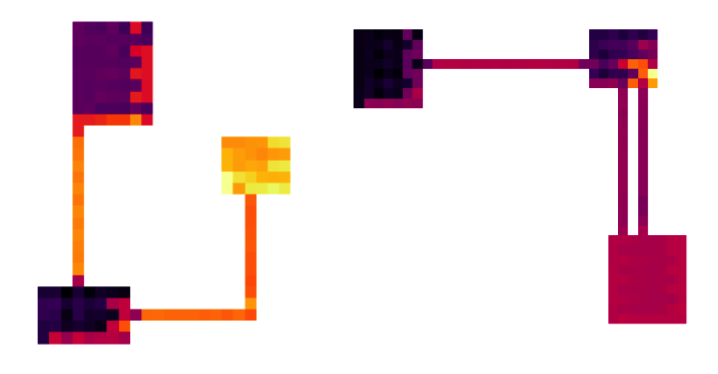}}
\caption{State encoding reward}
\label{Fig:handshapeSame}
\end{subfigure}
\caption{Qualitative results the Maze domain. To obtain these results, we spawned the agent in each location of the maze, and then produced a heat map of the rewards received.}\label{fig:rewards}
\end{figure}
We now discuss results for training the CDPRs for each task and running RL with the learned rewards.

\subsection{Maze results}
\begin{figure}[h]
\centering
\begin{subfigure}[b]{6.5cm}
\includegraphics[height=4.5cm]{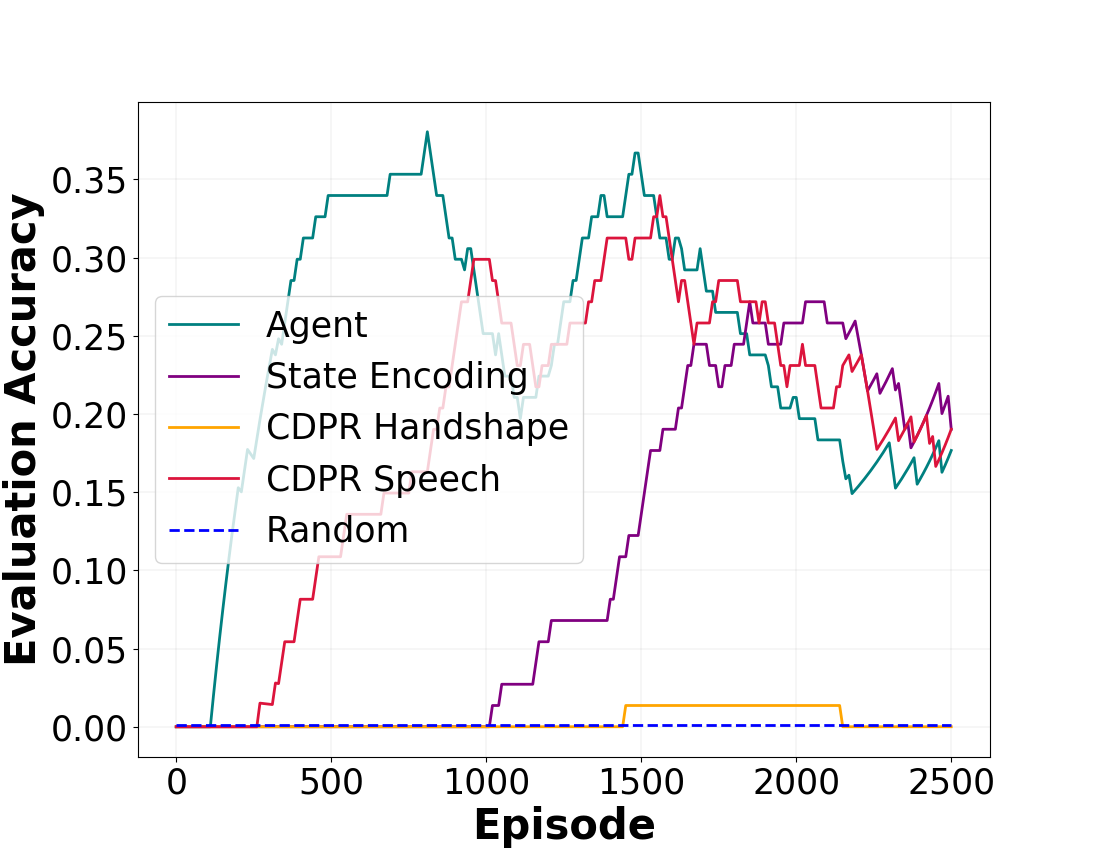}
\caption{Task 1}
\label{fig:mazeresults0}
\end{subfigure}
\hspace{.35cm}
\begin{subfigure}[b]{6.5cm}
\includegraphics[height=4.5cm]{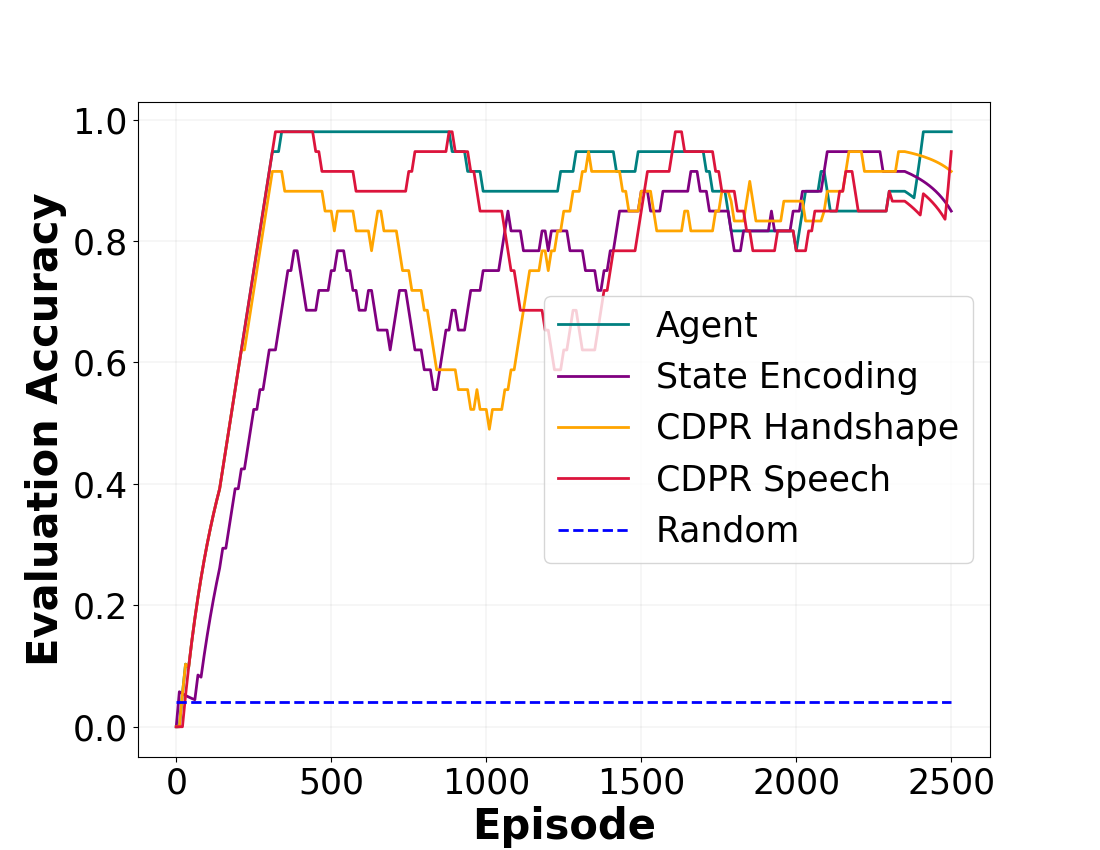}
\caption{Task 2}
\label{fig:mazeresults1}
\end{subfigure}
\caption{Smoothed results for running RL in the Maze domain. We ran RL with the goal specifications described in Figure~\ref{fig:MazeDomain}. The evaluation accuracy measures the percentage of time the agent spent in the correct room for each episode.}
\label{fig:mazeresults}
\end{figure}
We first give qualitative results for the Maze task. Figure~\ref{fig:rewards} shows a heat map of the rewards obtained from each location the agent could be spawned in. We do not show results for the standard reward, since this would just give rewards of 1 in the correct room and 0 otherwise. It is clear that CDPRs have learned to accurately reward the correct rooms for both the handshape and speech goal specifications. The largest reward values are given when the agent is in the correct room. We report the rewards for the feature encoding reward only for comparison, as these rewards are not directly learned. It is clear that the correct room will receive higher rewards than the rest, since the reward will be 0 when the state and goal are equal.

The next result we report is the GRA, shown in Table~\ref{table:retrieval}. Again, we show the encoding results for completeness, but the rewards will always yield a goal retrieval accuracy of 1. The handshape and speech CDPRs achieve high accuracy for retrieving the correct goal, demonstrating the accuracy of the approach and that it can generalize across new, unseen, goals.

Finally, we report the results for using the CDPRs with deep RL, as shown in Figure~\ref{fig:mazeresults}. We were interested in finding not only if the agent reached the goal, but also if it remained there, since this domain did not have terminal states. The desired goals were for the agent to reach the blue and green rooms, respectively, in the rooms displayed in Figure~\ref{fig:MazeDomain}. The CDPR with the spoken command was able to perform well on both tasks. Interestingly, the CDPR with the handshape only performed well one one task. This may be surprising, since this goal representation achieved a high GRA. It is clear that the GRA should not be the only metric for evaluating rewards. Rather, we should  utilize multiple evaluation metrics, as incorrectly specified rewards have been known to lead to strange behavior. We have one explanation for why the handshape goal representation does not work for the first goal specification. If we again study the qualitative results from Figure~\ref{fig:rewards}, we observe that the handshape CDPR actually gives intermediate rewards in the hallways. This may lead to suboptimal policies, so even though the GRA is high, these intermediate rewards may lead to locally suboptimal policies. The feature encoding network also gives intermediate rewards, but these rewards will not introduce positive cycles since none of the rewards are greater than 0. Future work then should ensure we train on both goal images and random states obtained from the domain to ensure ``false positives'' such as this do not occur.

\subsection{Music results}
\begin{figure}[h]
\centering
\begin{subfigure}[b]{6.5cm}
\includegraphics[height=4.5cm]{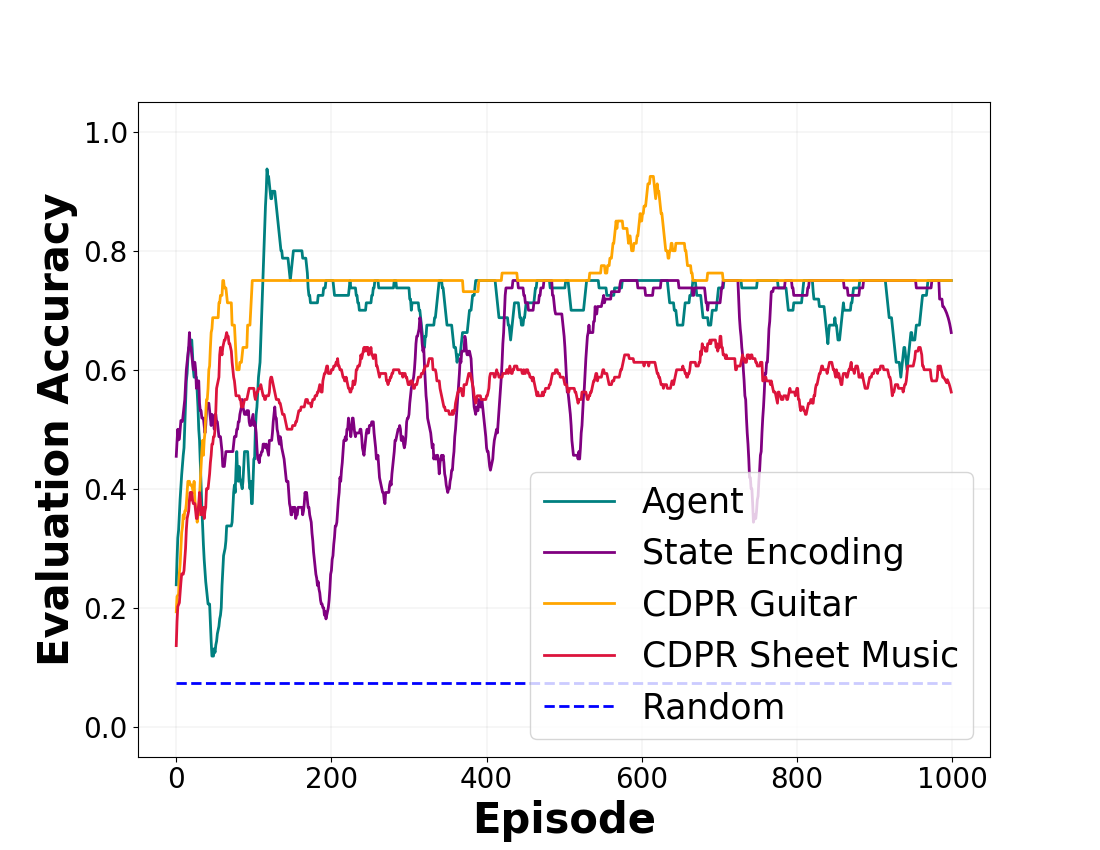}
\caption{Task $1$}
\label{Fig:MazeDomain}
\end{subfigure}
\hspace{.35cm}
\begin{subfigure}[b]{6.5cm}
\includegraphics[height=4.5cm]{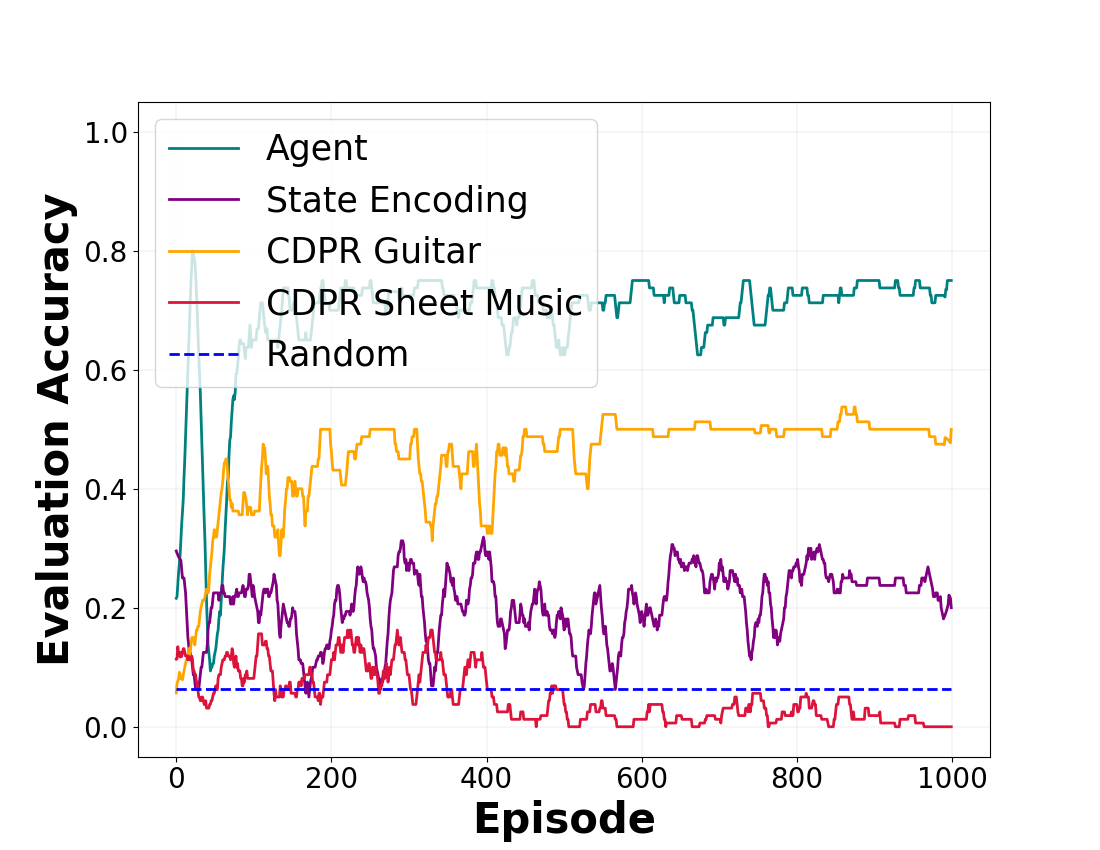}
\caption{Task $2$}
\label{Fig:MusicDomain}
\end{subfigure}
\caption{Smoothed results for running RL in the Music domain. We ran RL with the goal specifications described in Figure~\ref{fig:MusicDomain}. The evaluation accuracy measures the percentage of notes the agent played correctly at the end of each episode.}
\label{fig:musicresults}
\end{figure}
We now report results for the music domain. We do not have qualitative results for this task as the rewards are more difficult to visualize. The first result we report then is the GRA, shown in Table~\ref{table:retrieval}. The guitar and sheet specifications also achieved high accuracy, but the results are clearly not as good as the Maze results. Just as in classification problems, we will need to determine how much error we are willing to tolerate to avoid providing solutions for every problem. Nevertheless, future work will aim to develop more sophisticated architectures that produce higher GRAs.

We use the RL results to further examine the correctness of the CDPRs, as shown in Figure~\ref{fig:musicresults}. The desired goals were for the agent to play the songs depicted in Figure~\ref{fig:MusicDomain}: \{(`a', W), (`b', W), (`b', H), (`c', H), (`d', W)\} and \{(`f', W), (`a', H), (`e', H), (`a', H), (`c', H), (`e', W)\}, where $W$ and $H$ are whole and half notes, respectively. Both the guitar and sheet music goal specifications yielded results with high accuracy on the first song, each achieving accuracies of up to $.75$. The second song was slightly harder, as it had more notes. Both specifications achieved lower accuracy on the second song. In fact, the sheet music encoding learned a policy worse than random. The correspondence for this goal representation is likely more difficult than the guitar representation, which we see some evidence of in Figure~\ref{table:retrieval}. As we make improvements to the GRAs, we also expect the accuracy in RL to improve as well.

\section{Discussion and conclusion}
\label{sec:conclusion}
Our results indicate that we can accurately learn general reward functions specified through CDPRs. As discussed in the related work section, there is clearly evidence of momentum for this area of research. Further research will entail improving the accuracy of the CDPRs and their performance on RL tasks. Additionally, our initial work focused on determining that CDPRs could effectively be used to represent cross-domain goal representations. Future work will aim to utilize real-world samples of goals, perhaps obtained from crowd-sourcing. As such, it will be necessary to improve the sample complexity required to learn the CDPRs, which we did not examine in this study.

In summary, we have shown how goals can be defined in alternative environments than the agents. We introduced a general reward function, CDPRs, and showed that our approach could achieve strong performance in tasks specified through cross-domain goal specifications.

\small
\bibliographystyle{abbrv}
\bibliography{references}

\end{document}